\documentclass[10pt,journal,compsoc]{IEEEtran}
\RequirePackage{etex}

\usepackage{amsmath,amssymb,amsfonts,amsthm}
\usepackage{graphicx}
\usepackage{textcomp}
\usepackage{xcolor}
\usepackage{booktabs}
\usepackage{threeparttable}
\usepackage{wasysym}
\usepackage[ruled,vlined,linesnumbered, noend]{algorithm2e}
\usepackage[lambda,advantage,operators,sets,adversary,landau,probability,notions,logic,ff,mm,primitives,events,complexity,asymptotics,keys]{cryptocode}
\usepackage{tcolorbox}
\usepackage{balance}

\ifCLASSOPTIONcompsoc
  \usepackage[nocompress]{cite}
  \usepackage[caption=false,font=footnotesize,labelfont=sf,textfont=sf]{subfig}
\else
  \usepackage{cite}
  \usepackage[caption=false,font=footnotesize]{subfig}
\fi

\usepackage{pifont}
\newcommand{\cmark}{\ding{51}}%
\newcommand{\xmark}{\ding{55}}%

\begin{document} 

\title{NN-EMD: Efficiently Training \textit{Neural Networks} using \textit{Encrypted Multi-sourced Datasets}}

\author{
    Runhua~Xu, ~\IEEEmembership{Member,~IEEE}
    James~Joshi,~\IEEEmembership{Senior Member,~IEEE} and
    Chao~Li
    
    \IEEEcompsocitemizethanks{
    	\IEEEcompsocthanksitem R. Xu is now with IBM Research, San Jose, CA 95120, while most of the work was done when he was affiliated with the University of Pittsburgh. J. Joshi is with the School of Computing and Information, University of Pittsburgh, PA, USA, 15260. C. Li is with Beijing Key Laboratory of Security and Privacy in Intelligent Transportation, Beijing Jiaotong University, Beijing, China, 100044. \protect\\
    	Corresponding author: R. Xu and C. Li \protect\\
		Emails: runhua@ibm.com, jjoshi@pitt.edu,li.chao@bjtu.edu.cn
	}
}

\IEEEtitleabstractindextext{
\begin{abstract}
Training complex neural network models using third-party cloud-based infrastructure among multiple data sources is a promising approach among existing machine learning solutions.
However, privacy concerns of large-scale data collections and recent regulations have restricted the availability and use of privacy sensitive data in the third-party infrastructure. 
To address such privacy issues, a promising emerging approach is to train a neural network model over an encrypted dataset. 
Specifically, the model training process can be outsourced to a third party such as a cloud service that is backed by significant computing power, while the encrypted training data keeps the data confidential from the third party.
Compared to training a traditional machine learning model over encrypted data, however, it is extremely challenging to train a deep neural network (DNN) model over encrypted data for two reasons: first, it requires large-scale computation over huge datasets; second, the existing solutions for computation over encrypted data, such as using homomorphic encryption, is inefficient. Further, for enhanced performance of a DNN model, we also need to use huge training datasets composed of data from  multiple data sources that may not have pre-established trust relationships among each other.
We propose a novel framework, \textit{NN-EMD}, to train DNN over \textit{encrypted multiple datasets} collected from multiple sources. Toward this, we propose a set of secure computation protocols using hybrid functional encryption schemes.
We evaluate our framework for performance with regards to the training time and model accuracy on the MNIST datasets.
We show that compared to other existing frameworks, our proposed \textit{NN-EMD} framework can significantly reduce the training time, while providing comparable model accuracy and privacy guarantees as well as supporting multiple data sources.
Furthermore, the depth and complexity of neural networks do not affect the training time despite introducing a privacy-preserving \textit{NN-EMD} setting.
\end{abstract}

}

\maketitle

\thispagestyle{plain}
\pagestyle{plain}

\section{Introduction}
Deep neural networks (DNN), also known as deep learning, have been increasingly used in many fields such as computer vision, natural language processing, and speech/audio recognition, \cite{lecun2015deep}.
Such DNN-based solutions usually consist of two phases: the \textit{training} phase and the \textit{inference} phase. 
In the \textit{training} phase, a well-designed neural network is provided as input a training dataset and an appropriate optimization algorithm to generate optimal parameters for the neural network; then, in the \textit{inference} phase, the generated model (i.e., optimal parameters) is used for inference tasks, namely, predicting a label for an input sample.

One of the critically needed components in DNN-based applications is a powerful computing infrastructure with higher performance CPUs and GPUs, larger memory storage, etc., \cite{goodfellow2016deep}.
The volume of training data is another critical component.
For instance, existing commercial Machine Learning (ML) service providers such as Google, Microsoft, and IBM have devoted significant efforts toward building \textit{infrastructure as a service} (IaaS) platforms for clients that do not have such powerful computing resources.
The clients can employ these ML-related IaaS to manage a large-scale dataset collected from multiple data sources and train their models and provide prediction services to their customers.
Thus, training a DNN model using third-party IaaS among multiple data sources becomes a promising approach in the DNN-based application ecosystem. 

However, in many scenarios, the training data used for DNN are highly privacy-sensitive in nature.
Recent data breach incidents have increased the privacy concerns related to the large-scale collection and use of personal data \cite{rosati2019social, vemprala2019social}. 
Moreover, recent regulations such as the European General Data Protection Regulation (GDPR), California Consumer Privacy Act, Cybersecurity Law of China, etc., restrict the availability and use of privacy sensitive data.
Such privacy concerns of users, and the requirements imposed by regulations pose a significant challenge for the deployment of DNN solutions. 

\begin{table*}[!t]
    \centering
    \caption{Comparison of representative privacy-preserving approaches in deep neural networks}
    \begin{threeparttable}
    \begin{tabular}{lccccl}
    \toprule
        Proposed Work    & Training\tnote{$\triangleright$}   & Prediction\tnote{$\triangleright$}  & Privacy Target  & Multiple Data Source &  Underlying Technique    \\
    \midrule
        Shokri et al. \cite{shokri2015privacy}    & \cmark & \xmark & partial model & horizontal & Distributed selective SGD \\
        Abadi et al.\cite{abadi2016deep} & \cmark & \xmark & final model & no & Centralized differential privacy \\
        SecureML\cite{mohassel2017secureml} & \cmark &\cmark & training/inference data & no & Customized SMC \\
        DeepSecure\cite{rouhani2018deepsecure}  & \xmark &\cmark & inference data & no & General SMC (Garbled Circuits)\\
        CryptoNets\cite{gilad2016cryptonets},\cite{hesamifard2017cryptodl, chabanne2017privacy,jiang2018secure} & \xmark & \cmark & inference data & no & Homomorphic Encryption \\
        Nandakumar et al.\cite{nandakumar2019towards} & \cmark &\cmark & training/inference data & no & Homomorphic encryption \\
        CryptoNN\cite{xu2019cryptonn} & \cmark &\cmark & training/inference data & no mention & Functional encryption \\
        NN-EMD (our work) & \cmark &\cmark & training/inference data & horizontal/vertical & Hybrid functional encryption \\
    \bottomrule
    \end{tabular}
    \begin{tablenotes}
        \footnotesize
        \item[$\triangleright$] The symbols in \textit{training} and \textit{prediction} columns indicate that whether the proposed approach is applicable in training phase, prediction phrase or both.
    \end{tablenotes}
    \end{threeparttable}
    \label{tab:comparsion_approach}
\end{table*}

To address the aforementioned challenges,
several approaches have been proposed in the literature for building privacy-preserving ML systems. These approaches include: 
(\romannumeral1) applying privacy-preserving mechanisms such as \textit{differential privacy} to limit the disclosure of private information before outsourcing a dataset to a third party to train a DNN model \cite{abadi2016deep}; 
(\romannumeral2) employing new DNN architectures such as \textit{federated learning} where each participant trains a model locally and exchanges only the model parameters with the coordinating server \cite{shokri2015privacy}; and
(\romannumeral3) utilizing existing general \textit{secure multi-party computation} techniques or other encryption schemes (e.g., homomorphic encryption) and protocols to protect the input training data while training a DNN model \cite{mohassel2017secureml, rouhani2018deepsecure, gilad2016cryptonets, nandakumar2019towards, xu2019cryptonn}.

In Table \ref{tab:comparsion_approach}, we summarize existing representative privacy-preserving approaches used by deep learning systems.
Existing solutions such as a \textit{federated learning} approach and those based on \textit{differential privacy} cannot provide strong privacy guarantees because of \textit{inference attacks}, as demonstrated in the literature \cite{fredrikson2015model, shokri2017membership, nasr2019comprehensive}.
The general \textit{secure multi-party computation} (garbled circuits based) approaches, such as those proposed in \cite{mohassel2017secureml, rouhani2018deepsecure}, have a limitation with regards to large volumes of encrypted data that need to be transferred during the execution of the associated secure protocols.
Except for the recently proposed solutions such as in \textit{CryptoNN} \cite{xu2019cryptonn} and \cite{nandakumar2019towards, gilad2016cryptonets}, most of these SMC approaches only address privacy issues in the \textit{inference} phase rather than in the \textit{training} phase; this is mainly due to the efficiency challenges related to both computation and communication.

Furthermore, none of the existing solutions consider the fact that training data may be coming from multiple data sources distributed horizontally or vertically. 
The training dataset may have different composition cases; it may include data from multiple data sources, where: (\romannumeral1) each data source provides a dataset that includes all the features; (\romannumeral2) each one provides a dataset that has only a subset of the features, but, collectively these datasets cover the complete set of features; or (\romannumeral3) it is a hybrid of  (\romannumeral1) and (\romannumeral2).
Even though existing \textit{CryptoNN} \cite{xu2019cryptonn} support case (\romannumeral1), cases (\romannumeral2) and (\romannumeral3) still pose a huge challenge when considering privacy-preserving training of a neural network model.

In this paper, we propose a framework to train a \textit{Neural Network over Encrypted Multi-sourced Datasets} (NN-EMD). That is, \textit{NN-EMD} trains a neural network using a dataset that is composed of independently encrypted datasets from many different sources.
Each data source may provide its encrypted data that may include a \textit{complete set of features} or only a \textit{subset of features}. 
The goal here is to provide a strong privacy guarantee, while training a DNN model more efficiently as compared to the most recently proposed solutions, namely, those in \cite{xu2019cryptonn, nandakumar2019towards}. 
The most closely related work \textit{CryptoNN} \cite{xu2019cryptonn} and our \textit{NN-EMD} both use functional encryption (FE) to address the problem of computing over encrypted data.
To tackle the challenge in the aforementioned three cases, in essence, \textit{NN-EMD} proposes a new method (i.e., hybrid FE solution) instead of the simple FE solution as adopted in \textit{CryptoNN}.
The implementation and experiments of \textit{NN-EMD} are also different from \textit{CryptoNN}, which will be discussed later.
To the best of our knowledge, \textit{NN-EMD} is the first efficient and more practical approach for training a DNN over a set of encrypted/private multi-sourced datasets.

Specifically, to tackle the challenges of secure computation between a \textit{server}, and a \textit{client pool} (data sources) contributing multiple datasets that are composed in various ways, we first propose two non-interactive secure computation protocols between a \textit{server} and the \textit{client pool}, namely, a \textit{secure two-party horizontally partitioned computation} (\textit{S2PHC}) and a \textit{secure two-party vertically partitioned computation} (\textit{S2PVC}) protocols.
We construct these two protocols by using two types of \textit{functional encryption} schemes. \textit{NN-EMD} uses these as the building blocks; in particular,
\textit{S2PHC} and \textit{S2PVC} are used in each training iteration according to varying data composition cases.
Furthermore, original \textit{NN-EMD} only protects the raw input data in the case where the client has limited computing power and the server takes all training computation tasks. 
\textit{NN-EMD} is also able to enhance the privacy guarantee by integrating other privacy-preserving DNN approaches such as \textit{SplitNN} \cite{vepakomma2018split} and federated learning \cite{konevcny2016federated}.
We also propose to integrate the \textit{SplitNN} technique to provide a stronger privacy guarantee.

We also implemented an \textit{NN-EMD} system that can be deployed in a real cloud environment to support training neural networks using a set of encrypted datasets from multiple sources that are independently pre-processed locally.
Each data source can independently encrypt its data samples with complete or partial features, or incomplete data samples with partial features. 
Based on the collected encrypted (incomplete) data samples, the remotely deployed \textit{NN-EMD} system is able to train a global model.

We analyze the security and privacy properties of our proposed \textit{NN-EMD} approach and show that it satisfies the security and privacy goals.
We evaluate the performance of \textit{NN-EMD} with regards to training time, local pre-processing time and model accuracy.
The experimental results on the MNIST dataset show that our proposed \textit{NN-EMD} approach achieves significant efficiency improvements by reducing training time by more than 90\% compared to that of the best existing homomorphic encryption based approach while achieving the comparable model accuracy and privacy guarantees in basic \textit{NN-EMD} setting.
We also explore privacy and efficiency trade-off in the \textit{SplitNN} integrated \textit{NN-EMD} setting.
Furthermore, the depth and complexity of neural networks do not affect the training time despite introducing a privacy-preserving \textit{NN-EMD} setting.

\noindent\textbf{Organization}. 
In Section \ref{sec:background}, we introduce the background and preliminaries.
We propose \textit{NN-EMD} in Section \ref{sec:nn-emd}, the associated threat model in Section \ref{sec:threat_model}, the underlying secure computation approaches in Section \ref{sec:sc}, and the details of the framework in Section \ref{sec:nn-emd_specific}.
The security and privacy analysis is presented in Section \ref{sec:sp} and the evaluation is presented in Section \ref{sec:evaluation}.
We discuss the related work in Section \ref{sec:related_work} and conclude the paper in Section \ref{sec:conclusion}.

\section{Background and Preliminaries}
\label{sec:background}

\subsection{Motivation}
Deep neural networks (DNN) are increasingly being developed for many application domains; but in many applications training data is highly privacy-sensitive, and hence, needs to be protected. Further, they need significant computational resources.
For instance, a DNN-based system for breast cancer screening can provide much more effective, efficient, and patient-centric breast cancer screening support than ever before \cite{Harvey2019}.
However, some small clinics may not be able to train a breast cancer screening ML model based on their collected patients' healthcare records because of a lack of adequate computing power and ML expertise.
So, as an alternative, they can use a commercial cloud service that can provide the required computational infrastructure or neural network architecture; but such outsourced computing is not desirable without employing appropriate privacy-preserving techniques that can guarantee users' or regulatory privacy protection requirements.
Furthermore, in some cases, the training data is vertically partitioned among various clinics. For instance, one clinic may have the blood test report while another clinic may own the x-ray images.

This paper focuses on a DNN approach that uses a client-server architecture with two parties: (\romannumeral1) the \textit{cloud service provider} (server) with powerful computational infrastructure that can be employed for training a DNN model; and (\romannumeral2) the \textit{client pool} (data sources) that have privacy-sensitive datasets and need to build a DNN model based on these training datasets without leaking private information.

Such privacy-preserving DNN needs novel secure computation protocols to support efficient computation and interactions between the client pool and the server, while offering strong privacy guarantees.
Existing general \textit{secure multi-party computation} (SMC) solutions (i.e., garbled circuits) have limitations because they need to perform several rounds of communication involving transmission of large volumes of intermediate data. 
Using these techniques for DNN is cost prohibitive because of the huge volumes of training data needed.
Cryptography-based solutions (e.g., homomorphic encryption-based SMC) also has computational efficiency problem.

To the best of our knowledge, the approach proposed by Bost et al. in \cite{bost2015machine} is the first work that supports both training and predictive analysis over encrypted data. 
Their approach achieves this by integrating several crypto schemes, i.e., Quadratic Residuosity cryptosystem, Paillier cryptosystem, and homomorphic encryption, with secure protocols designed for them.
However, their approach only supports limited types of basic ML models such as naïve bayes, decision trees and support vector machine, but not DNNs.
Most recently, approaches proposed by Nandakumar et al. in \cite{nandakumar2019towards}, and Xu et al. in \cite{xu2019cryptonn} are the only ones that support training neural networks over encrypted data; their approaches use homomorphic and functional encryption, respectively.
Insight from these two approaches indicates that the crypto-based secure computing techniques are promising for the training phase of a DNN model.
However, there are two key challenges toward achieving effective and efficient training of neural networks over encrypted datasets that we address in this paper: (\romannumeral1) \textbf{\textit{Efficiency of Training Process}}: The existing secure computing protocols are not efficient, as mentioned above. For instance, with optimized approaches (e.g., multiple threads, training data distillation) in \cite{nandakumar2019towards}, training time for one mini-batch, with 60 samples, is around 40 minutes because the computation of each layer of neural networks is over ciphertext. This indicates that training time in the case of a larger volume of training data will be significantly higher.
(\romannumeral2) \textbf{\textit{Multiple Data Sources}}. There is a lack of consideration of real, complex datasets composed of horizontally and vertically partitioned datasets coming from multiple data sources.
Meanwhile, the training techniques also provide strong confidentiality-level privacy guarantees.

\subsection{Functional Encryption}
\label{sec:fe}
In this paper, we use functional encryption to construct our secure computing protocols instead of homomorphic encryption that has been employed by most of the existing privacy-preserving machine learning approaches.

Generally, functional encryption (FE) belongs to a public-key encryption family \cite{lewko2010fully, boneh2011functional}, where the decrypting party can be issued a secret key, also known as a \textit{functionally derived key}, to allow it to learn the result of a function over a ciphertext without leaking the corresponding plaintext.
To construct functional encryption schemes for general functionality, most of the recently proposed approaches such as those in \cite{goldwasser2014multi,boneh2015semantically,waters2015punctured,garg2016candidate,carmer20175gen, lewi20165gen} only focus on the theoretical feasibility or the existence of functionalities, and not on the computational efficiency issues.
These schemes rely on strong primitives such as indistinguishable obfuscation or multilinear maps that are prohibitively inefficient \cite{kim2018function}.

As most underlying computational operations of the training and inference phases of a DNN can be classified as matrix multiplications, or more precisely, vector inner-products, we employ \textit{functional encryption for inner-product} (FEIP) scheme instead of the general functional encryption scheme that provides general functionality at the expense of inefficiency.
To be specific, we adopt two kinds of FEIP schemes: \textit{single-input FEIP} and \textit{multi-input FEIP}.
The security of both these schemes is based on the decisional Diffie-Hellman (DDH) assumption.

\noindent\textit{Single-input FEIP}. 
We adopt the single-input FEIP (SI-FEIP) construction proposed in \cite{abdalla2015simple}. 
In SI-FEIP scheme, the supported function is described as 
$f_{\text{SIIP}}(\pmb{x},\pmb{y}) = \sum^{\eta}_{i=1}(x_{i}y_{i}) \;\;\text{s.t.}\;\; |\pmb{x}|=|\pmb{y}|=\eta, $
where $\pmb{x}$ and $\pmb{y}$ are two vectors of length $\eta$, from different parties. 
The SI-FEIP scheme $\mathcal{F}_{\text{S}}$ includes four algorithms: \textit{Setup, SKGenerate, Encrypt, Decrypt}, and is defined as $\mathcal{F}_{\text{S}}=(\mathcal{F}_{\text{S}}.S, \mathcal{F}_{\text{S}}.K, \mathcal{F}_{\text{S}}.E, \mathcal{F}_{\text{S}}.D)$ in the rest of the paper.

\noindent\textit{Multi-input FEIP}.
We employ the multi-input FEIP (MI-FEIP) construction derived from the work proposed in \cite{abdalla2018multi}. 
In the MI-FEIP scheme, the support function is defined as $ f_{\text{MIIP}}((\pmb{x}_{1}, \pmb{x}_{2}, ..., \pmb{x}_{n}),\pmb{y}) = \sum^{n}_{i=1}\sum^{\eta_{i}}_{j=1}(x_{ij}y_{\sum^{i-1}_{k=1}\eta_{k}+j})
\;\text{s.t.}\;\; |\pmb{x}_i|=\eta_{i}, |\pmb{y}| = \sum^{n}_{i=1}\eta_{i}, $
where $\pmb{x}_i$ and $\pmb{y}$ are vectors from different parties.
Accordingly, the MI-FEIP scheme $\mathcal{F}_{\text{M}}$ includes five algorithms: \textit{Setup, PKDistribute, SKGenerate, Encrypt, Decrypt}, and is defined as $\mathcal{F}_{\text{M}}=(\mathcal{F}_{\text{M}}.S, \mathcal{F}_{\text{M}}.PK, \mathcal{F}_{\text{M}}.SK, \mathcal{F}_{\text{M}}.E, \mathcal{F}_{\text{M}}.D)$ in the rest of the paper. 

Note that there exist three roles/entities in the SI-FEIP and MI-FEIP schemes: (\romannumeral1) an \textit{encryptor} that employs \textit{encrypt} algorithm to protect the sensitive data; (\romannumeral2) a \textit{decryptor} that uses \textit{decrypt} algorithm to acquire function result; (\romannumeral3) a third-party authority (TPA) that runs the \textit{setup} algorithm to initialize the cryptosystem and then runs \textit{PKDistribute} and \textit{SKGenerate} algorithms to provide key service for both \textit{encryptors} and \textit{decrytors}.

\subsection{Neural Networks}
\label{sec:pre:nn}

Deep learning models are typically achieved by DNN in a hierarchical and non-linear architecture consisting of multiple layers.
Each layer includes several neural units (a.k.a, neurons) to receive the data generated from its previous layer and outputs the processed data for its next layer.
Such a structure allows higher-level, abstract features to be represented as lower-level features through non-linear function computation at each layer.

Usually, a DNN includes three types of layers: one \textit{input layer}; one \textit{output layer}; and several \textit{hidden layers}.
In particular, the raw data is encoded properly and fed into the input layer.
Then, the features are abstracted and mapped from the raw data gradually from the first layer to the last layer via non-linear activation functions and iterative update (a.k.a, gradient descent optimization algorithm) until the convergence condition (e.g., the limited training time, the specified number of iterations, the expected training accuracy) is reached.
Such mapping abstractions, also known as the learned model or neural weights, can be used to perform the inference/predictive tasks.

Suppose that a DNN includes $L$ layers. 
The computation of layer $l$ can be represented as $\pmb{a}_{l}=f_{\text{act}}(\pmb{w}_{l}\cdot\pmb{a}_{l-1}),$
where $\pmb{a}_{l}$ is the output of layer $l$, $f_{\text{act}}$ is the activation function, $\pmb{w}_{l}$ is the weight matrix of layer $l$.
The goal of training a DNN is to learn optimal neural weights $\pmb{W}$ based on a given training dataset $\mathcal{D}=\{(\pmb{x}^{(1)}, y^{(1)}), ..., (\pmb{x}^{(n)}, y^{(n)})\}$ and a specified loss function $\mathcal{L}$.
Such a problem can be described as 
$
\argmin_{\pmb{W}} E_{\mathcal{D}}(\pmb{W}) = \argmin_{\pmb{W}} \frac{1}{n}\sum^{n}_{i=1}\mathcal{L}(y^{(i)}, \pmb{a}_{l}^{(i)})).
$
As a non-linear optimization problem, common solutions include gradient descent and its variants.
In our work, our proposed secure computation approaches support the computation over the matrix.
Thus, we use the well-known mini-batch stochastic gradient descent (SGD) algorithm \cite{bottou2010large, li2014efficient} as the optimization algorithm, where the secure computation is over a mini-batch for each iteration.

\section{The Proposed NN-EMD Framework}
\label{sec:nn-emd}

\begin{figure*}[!t]
    \centering
    \includegraphics[scale=0.55]{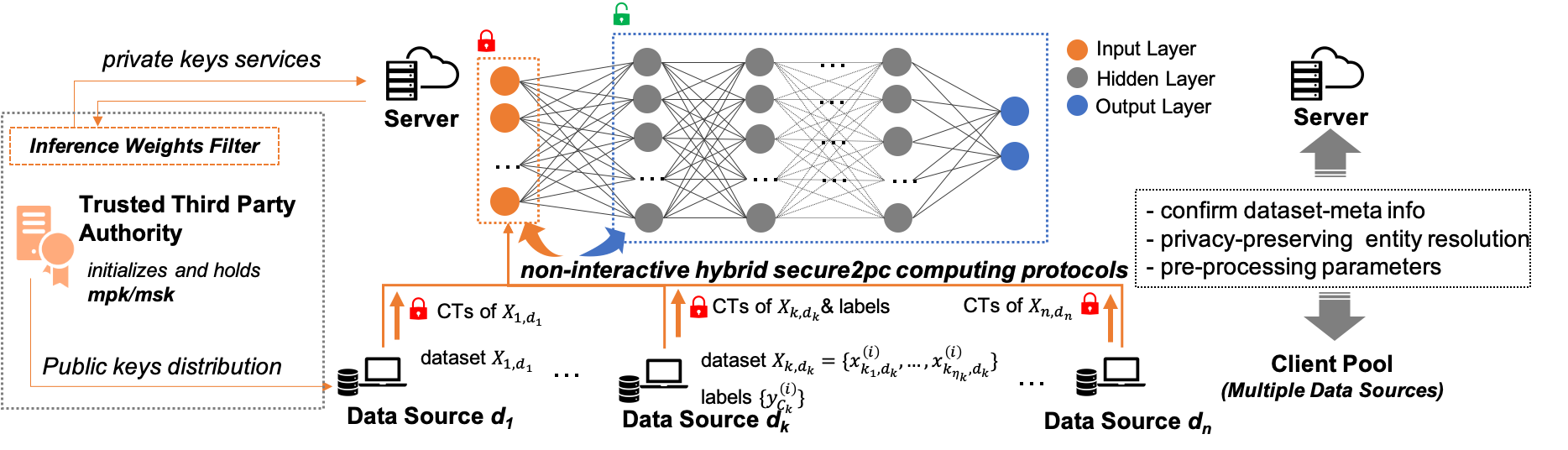}
    \caption{Framework overview of our proposed efficient approach for training deep neural networks from multiple data sources.}
    
    \label{fig:framework}
    
\end{figure*}

\subsection{Overview}

In \textit{NN-EMD}, we have the following three roles/entities: the \textit{client pool}, the \textit{server}, and a \textit{trusted third-party authority (TPA)}:
\begin{itemize}
    \item A \textit{client pool} of multiple \textit{data sources} that collaboratively contribute to the final training dataset composed of horizontally and vertically partitioned data, or a hybrid mix of the two. Each data source still keeps its data confidential from the rest.
    \item A \textit{server} responsible for training a DNN over a training dataset composed of multiple private datasets. 
    \item A trusted \textit{TPA} that initializes the underlying cryptosystems by setting secret credentials. Then, it distributes the associated public keys to data sources in the \textit{client pool} and the \textit{server}, and provides private key service to the \textit{server} during the training phase. \textit{Note that the TPA cannot acquire/access the encrypted training data}.
\end{itemize}

\figurename\;\ref{fig:framework} illustrates the essence of the \textit{NN-EMD} framework. 
Before the model training, the \textit{server} collects the meta information about the training datasets from the \textit{client pool} and then launches the privacy-preserving entity resolution mechanism with each data source if the final training dataset is a vertical composition of datasets from sources in the client pool.
We assume that for each data sample, there exists at least one data source having the label and only one data source's labels are enrolled in the training phase.
Meanwhile, both the \textit{client pool} and the \textit{server} acquire associated cryptographic keys from the \textit{TPA}.
Then, each data source in the \textit{client pool} pre-processes its data as required by the framework and outsources the encrypted data to the \textit{server}.
The \textit{server} starts to train the model by setting up the proper training hyperparameters, e.g., learning rate, number of iterations, and the total number of data sources, etc.
For instance, suppose we have two data sources $d_1$ and $d_2$ with datasets $\pmb{X}_{d_1}$ and $\pmb{X}_{d_2}$, respectively. In the case of horizontally partitioned dataset, $d_1$ and $d_2$ first prepare two types of ciphertext, $\enc(\pmb{X}_{d_1}), \enc(\pmb{X}^{\intercal}_{d_1})$ and $\enc(\pmb{X}_{d_2}), \enc(\pmb{X}^{\intercal}_{d_2})$ using the \textit{secure2pc} approach, respectively.
With received ciphertext, the server can launch the training using our proposed training algorithm based on \textit{secure2pc} approach. More detail will be presented in Section~\ref{sec:nn-emd_specific}.

\subsection{Threat Model and Assumptions}
\label{sec:threat_model}
We assume that there exists a \textit{trusted} TPA. 
This \textit{TPA} is an independent third-party that is widely trusted by all the \textit{data sources} in the \textit{client pool} and the \textit{server}. 
Note that it is also a common assumption in cryptosystems such as \cite{boneh2011functional, boneh2001identity, goyal2006attribute}.
The role of a trusted \textit{TPA} is similar to the role of a trusted \textit{certificate authority} in existing public key infrastructures. In this paper, we consider the following threat model:

\noindent (\romannumeral1) \textit{Honest-but-curious Server}; which is a common assumption in most of the existing approaches (\cite{bonawitz2017practical, xu2019cryptonn, nandakumar2019towards}). Here, the \textit{server} follows the instructions of a protocol or algorithm, but may try to learn private information by inspecting the collected encrypted dataset and decrypted functional results during the training phase.

\noindent (\romannumeral2) \textit{Curious and Colluding Data Sources}: In the \textit{client pool}, some of curious data sources may try to collude to infer any private information of other non-colluding data sources by inspecting their outsourced encrypted data.

\subsection{Secure Computation Approaches}
\label{sec:sc}

\begin{figure}[t]
    \centering
    \includegraphics[scale=0.5]{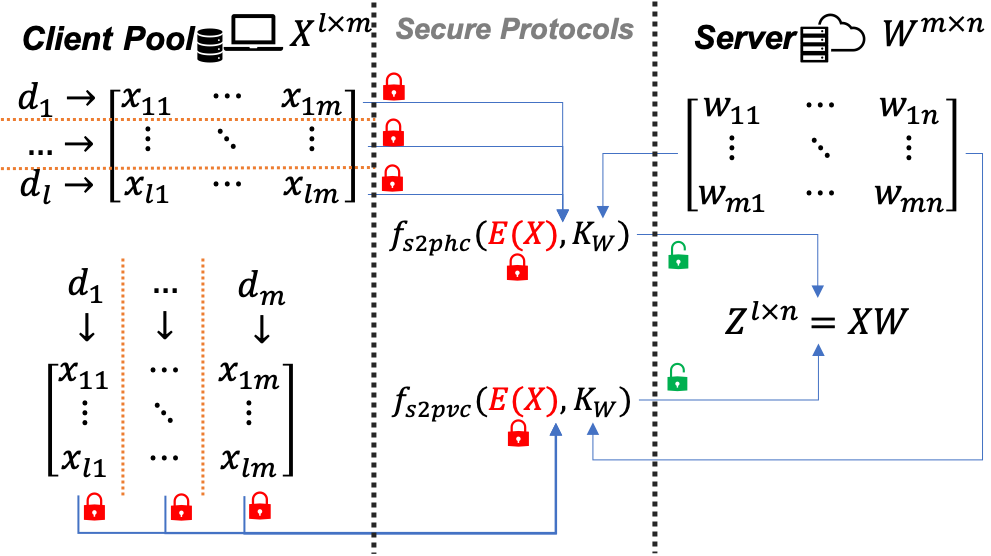}
    \caption{Illustration of secure two-party computation approaches between the client pool and the server.}
    
    \label{fig:overview_s2pc}
    
\end{figure}

\begin{figure*}[t]
    \centering
    \begin{tcolorbox}[arc=0.5mm, boxrule=0.05mm, center title, fontupper=\small, fontlower=\small, title=\textbf{Secure Two-party Horizontally Partitioned Computation Protocol }, title filled, colback=white, colbacktitle=white, coltitle=black]
    \textbf{Initialization and Key Services} \\
    \textit{$\Rightarrow$ \textit{TPA} initializes the system  as follows:}
    \begin{itemize}
        \item[-] initializes the single-input FEIP cryptosystem by generating a common public key and master private key, $pk_{\text{SI-FEIP},\text{com}}, msk_{\text{SI-FEIP}} \leftarrow \mathcal{F}_{S}.S(1^{\lambda}, 1^{\eta})$ by giving parameters $\lambda$ and $\eta$, where $\lambda$ is the security parameter indicating the bit length of security credentials, while $\eta$ denotes the maximum length of all possible input vectors of the inner-product function  $f_{\text{SIIP}}$ during the execution phase of the protocol.
        \item[-] initializes a private authenticated channel with the \textit{server} and the \textit{client pool}, respectively.
        \item[-] delivers the public key $pk_{\text{SI-FEIP}, \text{com}}$ and the parameter $\eta$ to both parties, namely, \textit{client pool} and \textit{server}. 
    \end{itemize}
    \textit{$\Rightarrow$ \textit{TPA} provides key services as follows:}
    \begin{itemize}
        \item[-] receives a functional private key request, and $\pmb{w}$ from the \textit{server}.
        \item[-] checks the $\pmb{w}$ to prevent potential inference attack by making sure $|\pmb{w}| \le \eta$ and non-zero elements of $\pmb{w}$ is less than the threshold $\tau$ using the weights filter module .
        \item[-] executes private key generation algorithm to generate private key $sk_{\text{SI-FEIP}, \pmb{w}} \leftarrow \mathcal{F}_{S}.K(msk, \pmb{w})$, and sends back the key via the private authenticated channel.
    \end{itemize}
    
    \textbf{Party: Client Pool}
    \textit{$\Rightarrow$ all data sources in the \textit{client pool} agree on an encoding precision $\epsilon_{\text{client}}$. For each data source $d_k \in \{d_1, ..., d_l\}$ in the client pool, each client in the pool executes the following steps:}
    \begin{itemize}
        \item[-] receives the public key $pk_{\text{com}}$ and $\eta$ from the \textit{TPA} and verifies the validity of $pk_{\text{com}}$. 
        \item[-] encodes elements in data from floating-point format $\pmb{X}_{\text{fp}}$ into integer format $\pmb{X}_{\text{int}}$ with encoding precision $\epsilon_{\text{client}}$.
        \item[-] counts the shape of the length of $\pmb{X}_{\text{int}} \rightarrow (s_{d_k}.r, s_{d_k}.c)$, and checks $s_{d_k}.c \le \eta$.
        \item[-] for each row $\pmb{x}_{i}$ of $\pmb{X}_{\text{int}}$, calls SI-FEIP encryption algorithm $ct_{\pmb{x}_i, d_k} \leftarrow \mathcal{F}_{S}.E(pk_{\text{com}}, \pmb{x}_i)$. 
        \item[-] if any above operations (assertion, verification, encoding, encryption) fails, abort.
        \item[-] sends all ciphertexts $\{ct_{\pmb{x}_1,d_k}, ..., ct_{\pmb{x}_{l},d_k}\}$ and parameters $\epsilon_{\text{client}}, (s_{d_k}.r, s_{d_k}.c)$ to the \textit{server}.
    \end{itemize}
    
    \textbf{Party: Server} \textit{$\Rightarrow$ the server executes the following steps:}
    \begin{itemize}
        \item[-] receives the public key $pk_{\text{SI-FEIP,com}}$ and $\eta$ from the \textit{TPA} and verifies the validity of $pk_{\text{SI-FEIP,com}}$.
        \item[-] collects ciphertexts $\textbf{ct} \leftarrow \{ct_{\pmb{x}_1,d_k},..., ct_{\pmb{x}_{l},d_k}\}$ and parameters $\epsilon_{\text{client}}$, $\{(s_{d_k}.r, s_{d_k}.c)\}$ from the \textit{client} party.
        \item[-] sets up the encoding precision $\epsilon_{\text{server}}$, and encodes each element in input weights from floating-point format $\pmb{W}_{\text{fp}}$  into integer format $\pmb{W}_{\text{int}}$. 
        \item[-] counts the shape of $\pmb{W}_{\text{int}} \rightarrow (s_{\text{server}}.r, s_{\text{server}}.c)$, and checks $\forall i,j, s_{\text{client}}.c \leftarrow s_{d_i}.c = s_{d_j}.c$ and $s_{\text{server}}.r = s_{\text{client}}.c \wedge s_{\text{server}}.r \le \eta$.
        \item[-] for each column $\pmb{w}_{i}$ of $\pmb{W}_{\text{int}}$, sends a function private key request to the \textit{TPA}, and collects the received private keys $\pmb{sk} \leftarrow \{sk_{f_{\text{SIIP}}, \pmb{w}_{1}}, ..., sk_{f_{\text{SIIP}}, \pmb{w}_{m}}\}$ with verification.
        \item[-] if all above operations (assertion, verification, encoding, encryption) fails, abort.
        \item[-] initializes a matrix $\pmb{Z}$ with shape $(|\pmb{ct}|, |\pmb{sk}|)$, and for each $i\in\{1, ...,|\pmb{ct}|\}$ and $j\in\{1, ...,|\pmb{sk}|\}$, calls decryption algorithm $w_{i,j} \leftarrow \mathcal{F}_{S}.D(pk_{\text{SI-FEIP,com}}, \pmb{ct}[i], \pmb{sk}[j], \pmb{w}_{j})$. 
        \item[-] decodes each element in $\pmb{Z}$ from integer format into floating-point format using $\epsilon_{\text{server}}$ and $\epsilon_{\text{client}}$.
    \end{itemize}
    
    \end{tcolorbox}
    
    \caption{Detailed description of non-interactive secure two-party horizontally partitioned computation protocol. Note that arrows indicate assignment operation, while the equal sign is a comparison operation.}
    \label{fig:protocol:s2phc}
\end{figure*}

\begin{figure*}[!t]
    \centering
    \begin{tcolorbox}[arc=0.5mm, boxrule=0.05mm, center title, fontupper=\small, fontlower=\small, title=\textbf{Secure Two-party Vertically Partitioned Computation Protocol}, title filled, colback=white, colbacktitle=white, coltitle=black]
    \textbf{Initialization and Key Services} \\
    \textit{$\Rightarrow$ TPA initializes the system as follows:}
    \begin{itemize}
        \item[-] initializes the multi-input FEIP crypto schemes by generating a common public key, master public key and private key $pk_{\text{MI-FEIP},\text{com}}, mpk_{\text{MI-FEIP}}, msk_{\text{MI-FEIP}} \leftarrow \mathcal{F}_{M}.S(1^{\lambda}, \vec{\pmb{\eta}}, n)$ by giving parameters $\lambda$ and $(\vec{\pmb{\eta}}, n)$, where $(\vec{\pmb{\eta}}, n)$ indicates the allowed $n$ maximum number of data sources where each data source has maximum input length represented as $\vec{\eta}$, during the computation execution of $f_{\text{MIIP}}$. 
        \item[-] assigns a identity $id_{d_k}$ for each registered data source $d_k$ in the \textit{client pool}.
        \item[-] initializes a private authenticated channel with the \textit{server} and the \textit{data sources}, respectively.
        \item[-] delivers $d_k$-associated $pk_{\text{MI-FEIP}, id_{d_k}} \leftarrow \mathcal{F}_{M}.PK(mpk_{\text{MI-FEIP}}, msk_{\text{MI-FEIP}}, id_{d_k})$, $\eta_{id_{d_k}} \leftarrow \vec{\eta}$ and the common public key $pk_{\text{MI-FEIP}, \text{com}}$ to each data source $id_{d_k}$, respectively.
        \item[-] delivers the common public key $pk_{\text{MI-FEIP}, \text{com}}, \vec{\pmb{\eta}}, n$ to the \textit{server}.
    \end{itemize}
    \textit{$\Rightarrow$ TPA provides key services:}
    \begin{itemize}
        \item[-] receives the request $\pmb{w}$ from the \textit{server}.
        \item[-] checks $\pmb{w}$ to prevent potential inference attack by checking that non-zero elements of $\pmb{w}$ is less than the threshold $\tau$ using weights filter module.
        \item[-] generates private key $sk_{\text{MI-FEIP}, \pmb{w}} \leftarrow \mathcal{F}_{M}.SK(mpk_{\text{MI-FEIP}}, msk_{\text{MI-FEIP}}, \pmb{w})$, and sends back the key via the private authenticated channel.
    \end{itemize}
    
    \textbf{Party: Client Pool}
    \textit{$\Rightarrow$ all data sources agree on an encoding precision $\epsilon_{\text{client}}$. For each data source $d_k \in \{d_1, ..., d_m\}$ in the client pool, each client executes the following steps:}
    \begin{itemize}
        \item[-] receives the public key $pk_{\text{MI-FEIP}, \text{com}}$, $pk_{\text{MI-FEIP}, id_{u_k}}$ and $\eta_{id_{u_k}}$ from the \textit{TPA} and verifies the validity of $pk_{\text{MI-FEIP}, \text{com}}$ and $pk_{\text{MI-FEIP}, id_{u_k}}$. 
        \item[-] encodes elements in data from floating-point format $\pmb{X}_{\text{fp}}$ into integer format $\pmb{X}_{\text{int}}$ with encoding precision $\epsilon_{\text{client}}$.
        \item[-] counts the shape of the length of $\pmb{X}_{\text{int}} \rightarrow (s_{d_k}.r, s_{d_k}.c)$, and checks $s_{d_k}.c \le \eta_{id_{u_k}}$.
        \item[-] for each row $\pmb{x}_{i}$ of $\pmb{X}_{\text{int}}$, calls MI-FEIP encryption $ct_{\pmb{x}_i, d_k} \leftarrow \mathcal{F}_{M}.E(pk_{\text{com}}, \pmb{x}_i)$. 
        \item[-] if any above operations (assertion, verification, encoding, encryption) fails, abort.
        \item[-] sends all ciphertexts $\{ct_{\pmb{x}_1,d_k}, ..., ct_{\pmb{x}_{l},d_k}\}$ and parameters $\epsilon_{\text{client}}, (s_{d_k}.r, s_{d_k}.c)$ to the \textit{server}.
    \end{itemize}
    
    \textbf{Party: Server} \textit{$\Rightarrow$ the server executes the following steps:}
    \begin{itemize}
        \item[-] receives the public key $pk_{\text{MI-FEIP},  \text{com}}, \vec{\pmb{\eta}}, n$ from the \textit{TPA} and verifies the validity of $pk_{\text{MI-FEIP}, \text{com}}$.
        \item[-] collects the ciphertexts $\textbf{ct} \leftarrow \{\{ct_{\pmb{x}_{i},d_1}\},...,\{ct_{\pmb{x}_{i},d_m}\}\}$ and parameters $\epsilon_{\text{client}}, \{(s_{d_k}.r, s_{d_k}.c)\}$ from the \textit{client pool}.
        \item[-] sets up the encoding precision $\epsilon_{\text{server}}$ and encodes each element in input weights from floating-point number $\pmb{W}_{\text{fp}}$ into integer number $\pmb{W}_{\text{int}}$. 
        \item[-] counts the shape of $\pmb{W}_{\text{int}} \rightarrow (s_{\text{server}}.r, s_{\text{server}}.c)$, and checks $\forall i,j, s_{d_i}.r = s_{d_j}.r$ and $s_{\text{server}}.r = \sum{s_{d_i}.c} \wedge s_{\text{server}}.r \le \sum\vec{\pmb{\eta}} \wedge |\pmb{ct}| < n$.
        \item[-] for each column $\pmb{w}_{i}$ of $\pmb{W}_{\text{int}}$, sends a function private key request to the \textit{TPA}, and collects the received keys $\pmb{sk} \leftarrow \{sk_{f_{\text{MIIP}}, \pmb{w}_{1}}, ..., sk_{f_{\text{MIIP}}, \pmb{w}_{m}}\}$ with verification.
        \item[-] if all above operations (assertion, verification, encoding, encryption) fails, abort.
        \item[-] re-organizes $\pmb{ct} \rightarrow \pmb{ct}^{'}$ by aggregating by $\pmb{ct}$ index. 
        \item[-] initializes a matrix $\pmb{Z}$ with $|\pmb{ct}^{'}|$ rows and $|\pmb{sk}|$ columns, and for each $i\in\{1, ...,|\pmb{ct}^{'}|\}$ and $j\in\{1, ...,|\pmb{sk}|\}$, and calls decryption algorithm $w_{i,j} \leftarrow \mathcal{F}_{M}.D(pk_{\text{com}}, \pmb{ct}[i], \pmb{sk}[j], \pmb{w}_{j})$. 
        \item[-] decodes each element in $\pmb{Z}$ from integer format into float point format using $\epsilon_{\text{server}}$ and $\epsilon_{\text{client}}$.
    \end{itemize}
    
    \end{tcolorbox}
    
    \caption{Detailed description of non-interactive secure two-party vertically partitioned computation protocol.}
    \label{fig:protocol:s2pvc}
\end{figure*}

Here, we present our proposed privacy-preserving secure computation approach between a \textit{server} and a \textit{client pool} (data sources).
To be specific, we propose two secure computation protocols, namely, \textit{secure two-party horizontally partitioned computation} protocol (\textit{S2PHC}, see \figurename\;\ref{fig:protocol:s2phc}) and \textit{secure two-party vertically partitioned computation} protocol (\textit{S2PVC}, see \figurename\;\ref{fig:protocol:s2pvc}). 
Both the protocols are non-interactive, secure two-party computation protocols, where there is no interaction among data sources in the \textit{client pool}, and not interaction between the \textit{server} and the \textit{client pool}; i.e., there only exists one-way communication from \textit{client pool} to the \textit{server}.

The difference between the two secure computation protocols is mainly with regards to how the input of computing function from \textit{client pool} are composed. 
Suppose that there is a secure computation task such as a matrix multiplication $\pmb{X}^{l\times m}\pmb{W}^{m\times l}$ between the \textit{client pool} and the \textit{server}, where the \textit{client pool} has the matrix $\pmb{X}^{l\times m}$ that is composed of data from different data sources $\{d_k\}$ and the \textit{server} has $\pmb{W}^{m\times l}$.
As shown in \figurename\;\ref{fig:overview_s2pc}, $\pmb{X}^{l\times m}$ represents horizontal or vertical composition of data from multiple sources.

\noindent\textbf{S2PHC}.
We present the detailed description of the \textit{non-interactive S2PHC} protocol in \figurename\;\ref{fig:protocol:s2phc}.
The protocol is built from the single-input functional encryption scheme.
Here, we suppose that each data source in the \textit{client pool} has the same column length related to $\pmb{X}$ as illustrated in \figurename\;\ref{fig:overview_s2pc}. 
It indicates that each data source owns complete features for each data sample, and those data samples constitute a training dataset.
Note that the \textit{S2PHC} protocol can be considered as an improvement of the secure matrix computation approach proposed in \cite{xu2019cryptonn} where the possibility of multiple horizontal data sources had been mentioned, but no theoretical analysis and practical implementation were presented. 
Unlike in \cite{xu2019cryptonn}, we present specific practical construction in our protocol with the experimental evaluation in Section \ref{sec:evaluation}.

\noindent\textbf{S2PVC}.
\figurename\;\ref{fig:protocol:s2pvc} shows the details of the \textit{S2PVC} protocol.
Here, we assume that each data source from the \textit{client pool} has the same row length with regards to $\pmb{X}$.
It means that each data source that owns partial features can provide the same size of data samples to constitute a training batch, while those partial features can compose the complete features.
The \textit{S2PVC} protocol is constructed using the multi-input functional encryption scheme as the key underlying scheme.

\subsection{NN-EMD Training}
\label{sec:nn-emd_specific}

Here, we present the details of our proposed \textit{NN-EMD} framework.
As mentioned above, \textit{NN-EMD} mainly includes two parties: the \textit{server} and the \textit{client pool}, and they use \textit{S2PHC} and \textit{S2PVC} protocols.
Suppose that there exists data sources $S_{d}$=$\{d_1, ..., d_m\}$, where each data source $d_k \in S_{d}$ has dataset $\pmb{X}_{d_k}$.
The goal of the \textit{NN-EMD} framework is to train a neural network model based on the dataset $\pmb{X}$ that is composed of $\{\pmb{X}_{d_1}, ..., \pmb{X}_{d_m}\}$ without leaking $\pmb{X}$ to the \textit{server}, and without disclosing $\pmb{X}_{d_i}$ to $d_j$ where $d_i, d_j \in S_d \wedge d_i\ne d_j$.
Such an assumption is common in existing vertical machine learning related literature, and also indicates there are no overlapping features among those data sources except for the identity feature used for the privacy-preserving entity resolution.

\begin{algorithm}[!t]
\SetAlgoLined
\caption{\textit{NN-EMD} Training Algorithm}
\label{alg:framework}
\footnotesize
\KwIn{secure parameter $1^{\lambda}$, functionality parameters $(\eta, \vec{\pmb{\eta}}, n)$, data sources $S_{d}=\{d_k\}$, each data source $d_{k}$ has dataset $\pmb{X}_{d_k}$.}
\KwOut{trained model $\pmb{W}$}
\SetKwProg{Fn}{function}{}{}
\SetKwProg{Pc}{party client}{}{}
\SetKwProg{Ps}{party server}{}{}
\SetKwRepeat{Do}{do}{while}%
initialize \textit{S2PHC} protocol by setting $(1^{\lambda}, \eta)$\;
initialize \textit{S2PVC} protocol by setting $(1^{\lambda}, \vec{\pmb{\eta}}, n)$\;
$p_{\text{batch}}, T_{d_k}\leftarrow$exchange meta-information of $\{\pmb{X}_{d_k}\}$ \;
\Pc{pre-process($\{\pmb{X}_{d_k}, p_{\text{batch}}, p_{T_{\pmb{X}_{d_k}}}\}$)}{
    \ForEach{$d_k \in S_{d_k}$}{
        \If{$T_{d_k}$=$T_f$}{
        \ForEach{mini batch $\pmb{X}_{d_k, \text{batch}} \in \pmb{X}_{d_k}$ }{
            $S_{\pmb{ct}_{\text{ff}}}\leftarrow$ \text{S2PHC}($d_k, \pmb{X}_{d_k, \text{batch}}$)\;
            $S_{\pmb{ct}_{\text{bp}}}\leftarrow$ \text{S2PHC}($d_k, \pmb{X}^{\intercal}_{d_k, \text{batch}}$)\;
        }
        }
        \Else{
            start entity resolution with shuffle\;
            \ForEach{mini batch $\pmb{X}_{d_k, \text{batch}} \in \pmb{X}_{d_k}$ }{
            $S_{\pmb{ct}_{\text{ff}}}\leftarrow$ \text{S2PVC}($d_k, \pmb{X}_{d_k, \text{batch}}$)\;
            $S_{\pmb{ct}_{\text{bp}}}\leftarrow$ \text{S2PHC}($d_k, \pmb{X}^{\intercal}_{d_k, \text{batch}}$)\;
        }
        }
        sends $S_{\pmb{ct}_{\text{ff}}}, S_{\pmb{ct}_{\text{bp}}}, T_{d_k}$ and $Y$ if $d_k$ has the label\;
    }
}
\Ps{training($\{S_{\pmb{ct}_{\text{ff}}}, S_{\pmb{ct}_{\text{bp}}}, T_{d_k}, \pmb{Y}\}$)}{
    $\pmb{W} \leftarrow$ initialize model weights\;
    \ForEach{iteration}{
        \ForEach{mini batch $\pmb{ct}_{\text{ff}} \in S_{\pmb{ct}_{\text{ff}}}, \pmb{ct}_{\text{bp}} \in S_{\pmb{ct}_{\text{bp}}}$}{
            \lIf{$T_{d_k}$=$T_p$}{$\pmb{A}_{1} \leftarrow \text{S2PVC}(\text{server}, \pmb{ct}_{\text{ff}})$}
            \lElse{$\pmb{A}_{1} \leftarrow \text{S2PHC}(\text{server}, \pmb{ct}_{\text{ff}}, \pmb{W}_{1})$}
            $\pmb{A}\leftarrow$ feed-forward($\pmb{A}_{1}, \pmb{W}$)\;
            $\pmb{\nabla}_{2,...,l},\pmb{\sigma}\leftarrow$ gradient compute( $\pmb{Y}, \pmb{W}, \pmb{A}$)\;
            $\pmb{\nabla}_{1}\leftarrow \text{S2PHC}(\text{server}, \pmb{ct}_{\text{bp}}, \pmb{\sigma})$\;
            $\pmb{W} \leftarrow \pmb{W} - \alpha\nabla$\;
        }
    }
    \Return $\pmb{W}$
}
\end{algorithm}

Algorithm \ref{alg:framework} illustrates how our proposed \textit{S2PHC} and \textit{S2PVC} protocols are integrated in the training process of a neural network model.
First, we initialize \textit{S2PHC} and \textit{S2PVC} protocols with proper security parameter $1^{\lambda}$ and function parameters $(\eta, \vec{\pmb{\eta}}, {n})$ as defined in Section \ref{sec:sc}.
Then, the \textit{server} acquires the basic meta-information of the training dataset from each source from the \textit{client pool}, and decides several training hyperparameters such as proper mini-batch size $p_{\text{batch}}$ and dataset type $T_{d_{k}}$ shared with each data source (lines 1-3).
Note that we define dataset types: $T_f$ and $T_p$ to indicate a dataset with a full or partial set of features corresponding horizontally and vertically partitioned datasets cases, respectively.

According to different compositions of final training data $\pmb{X}$, we propose three different training approaches: \textit{horizontally partitioned based training}, \textit{vertically based partitioned training}, and \textit{hybrid partitioned based training}.

\noindent\textbf{Horizontal Partitioning Based Training}.
This approach deals with the case where each data source's dataset has a full set of features needed in the training.
That is, $\pmb{X}$ is horizontally composed of $\{\pmb{X}_{d_1}, ..., \pmb{X}_{d_m}\}$.
In this case, each data source first divides its local dataset into several mini-batches according to the received batch parameter.
Then, for each mini-batch, the data source executes \textit{S2PHC} protocol twice with input mini-batch $\pmb{X}_{d_k, batch}$ and its transpose $\pmb{X}_{d_k, batch}^{\intercal}$, respectively.
The generated ciphertexts $S_{\pmb{ct}_{\text{ff}}}$ and $S_{\pmb{ct}_{\text{bp}}}$ are used in feed-forward computation and back-propagation computation in the training phase, respectively (lines 6-11).

On the server side, weights are randomly initialized for the model (line 22).
For each mini-batch iteration, \textit{S2PHC} protocol is executed with $S_{\pmb{ct}_{\text{ff}}}$ to support the secure computation that occurs between the input layer and the first hidden layer (line 25).
As the output is in plaintext, the normal feed-forward operations can be continued as in a normal neural network training phase (line 27).
In the back-propagation phase, the normal gradient computation can be done first from the last layer (line 28).
When it comes to the first layer, the \textit{server} executes the \textit{S2PHC} protocol with different ciphertext, namely, $S_{\pmb{ct}_{\text{bp}}}$ (line 29).
Finally, the weights are updated using the learning rate and current gradients (line 30) defined in Section \ref{sec:pre:nn}.

\noindent\textbf{Vertical Partitioning Based Training}.
This approach is for the case where each data source's dataset has a subset of features, however, these partial features collected from all the sources form the complete set of features; i.e., $\pmb{X}$ is vertically composed of $\{\pmb{X}_{d_1}, ..., \pmb{X}_{d_m}\}$.
Note that we assume that each $\pmb{X}_{d_k}$ has an identity column so that the privacy-preserving entity resolution mechanism can be executed; there are no overlapping features that will be used in the training. In this case, each data source starts with a privacy-preserving entity resolution mechanism with the \textit{server} that plays the role of a coordinator, similar to those in other approaches such as in \cite{schnell2011novel, ion2019deploying}.
Here, each data source sends the encoded identical features to the \textit{server} for entity matching. 
Then, the \textit{server} generates a proper permutation for each data source to re-order its local data.
As a result, a data source does not know which entity in its dataset has been enrolled in the training; and the \textit{server} still cannot learn the training dataset. As entity resolution is not the core contribution in our framework, we refer the reader to \cite{ion2019deploying} for more details.

Here, each data source generates $S_{\pmb{ct}_{\text{ff}}}$ by executing the \textit{S2PVC} with input $\pmb{X}_{d_k, batch}$, while generating $S_{\pmb{ct}_{\text{bp}}}$ by executing \textit{S2PHC} with input $\pmb{X}_{d_k, batch}^{\intercal}$ (lines 14-17).
The \textit{server} acquires the output of the first hidden layer by executing the \textit{S2PVC} protocol with corresponding $S_{\pmb{ct}_{\text{ff}}}$ (line 25).

\noindent\textbf{Hybrid Partitioning Based Training}.
Our \textit{NN-EMD} framework can also be naturally applied to the hybrid case where $\pmb{X}$ is composed of the data from multiple data sources using a mix of horizontal and vertical composition.
Algorithm \ref{alg:framework} is for processing the hybrid training case by integrating the horizontally partitioned based training approach with the vertically partitioned based training approach.

\noindent\textbf{Integration with SplitNN}.
If the data source may have computation power to training the partial DNN model, \textit{NN-EMD} also allows protecting more DNN layers. Such a setting is actually the integration of \textit{NN-EMD} with \textit{SplitNN} \cite{vepakomma2018split} framework. 
Here, \textit{NN-EMD} protects the splitting layer's output from the party side instead of protecting the raw data using the \textit{secure2pc} protocol.
Then, the server acquires the \textit{secure2pc} result to continue the feed-forward computation on the server-side.
In the backpropagation phase, the server computes corresponding gradients and passes the splitting layer gradients to the party for local gradients computation.

\noindent\textbf{Comparison with Existing Solutions}.
Here we briefly compare our \textit{NN-EMD} framework with \textit{CryptoNN} \cite{xu2019cryptonn} and the one in \cite{nandakumar2019towards}.
\textit{CryptoNN} is actually a special instance of our \textit{NN-EMD} framework in the horizontal partitioning based training setting.
Unlike those in \cite{xu2019cryptonn} \cite{nandakumar2019towards}, \textit{NN-EMD} does not protect the label information in the training dataset.
Actually, the encrypted label information in \textit{CryptoNN} framework can be easily inferred, while the design of encrypting label in \cite{nandakumar2019towards} is required by the adoption of underlying homomorphic encryption.
We argue that \textit{NN-EMD} satisfies the privacy requirements even though the label is exposed to the \textit{server}; we analyze this in Section \ref{sec:sp}.
In \cite{nandakumar2019towards}, all the outputs of each layer are still in ciphertext form. The output of the first hidden layer in \textit{NN-EMD} is in plaintext; because of which the training time does not increase as in \cite{nandakumar2019towards}.

Note that we do not present the \textit{inference} phase of the neural network model since the \textit{inference} can be viewed as one iteration of feed-forward computation in the training phase, as shown in Algorithm \ref{alg:framework}.

\section{Security and Privacy Analysis}
\label{sec:sp}


\subsection{Security of Underlying Cryptosystems}
\label{sec:sp:security}
\textit{S2PHC} and \textit{S2PVC} protocols are critical components of \textit{NN-EMD} framework that provides the basis for privacy guarantees.
As presented in Section \ref{sec:fe}, we add protocols to deliver the public keys and private keys generated by the \textit{TPA} on the originally proposed constructions of single-input and multi-input functional encryption schemes that we adopt for our proposed scheme.

For the formal proof of security of adopted functional encryption schemes we refer the readers to \cite{abdalla2015simple, abdalla2018multi}. In our adoption of these schemes, the added public key distribution and private key delivery methods are managed by the \textit{TPA}. This, however, does not affect the ordinal encryption and decryption constructions as compared to the originally proposed schemes.
With regards to the public-key setting in our framework with multiple data sources, each data source has its respective public key $pk_{\text{MI-FEIP}}$ and they all have a common public key $pk_{\text{SI-FEIP}}$.
Here, we analyze the possible security concern that a \textit{colluding} data source monitors or inspects the encrypted outsourced datasets from other data sources/clients.
Intuitively, such settings could enable the \textit{colluding} data sources in the client pool to infer the target encrypted data by iteratively encrypting its candidate data and then checking the ciphertext with target encrypted data as all sources share a common public key $pk_{\text{SI-FEIP}}$.
However, such an inference is prevented by the ciphertext indistinguishability property implied in the adopted functional encryption scheme \cite{abdalla2015simple, abdalla2018multi}.
For instance, for same input data $x$, with the same public key $pk_{\text{SI-FEIP}}$, the encrypted ciphertexts $c_1 = E_{pk_{\text{SI-FEIP}}}(x), c_2 = E_{pk_{\text{SI-FEIP}}}(x), ..., c_n = E_{pk_{\text{SI-FEIP}}}(x)$ are indistinguishable. 
That ciphertext indistinguishability is guaranteed by the IND-CPA security of SI-FEIP \cite{abdalla2015simple}.
Thus, there is still a non-negligible advantage for the attackers by increasing the number of colluding data sources to brute-force the encrypted data from the non-colluding data source \cite{abdalla2015simple}.
As a result, our framework can resist such a \textit{brute-force} attack by the \textit{colluding} data sources.

As mentioned earlier, the labels in our framework are not protected. 
We argue that such a design does not disclose the private information of the training data.
Essentially, in the binary classification task, the label is encoded into meaningless value such as using \{1,-1\} to represent positive and negative labels rather than using a meaningful/concrete label such as ``this x-ray image represents cancer''.
The \textit{server} can only learn group information of the encrypted data such as the information that $E_{\text{FE}}(\pmb{X}_{y=1})$ belongs to label $y=1$, but the \textit{server} cannot learn $\pmb{X}_{y=1}$, as it is protected by the cryptosystems, and what $y=1$ means.
The \textit{server} is also not able to launch the enrollment inference attack where the curious server tries to infer whether a target data is enrolled in the training or not, because the training data is encrypted via functional encryption.
In particular, the adopted FE schemes have the IND-CPA security guarantee, where the ciphertexts $c_i = E_{pk_{\text{SI-FEIP}}}(x), c_j = E_{pk_{\text{SI-FEIP}}}(x)$ of the same data $x$ is indistinguishable \cite{abdalla2015simple}.
Let us suppose the target of enrollment inference attack is $x_{\text{target}}$. The \textit{data source} encrypt $x_{\text{target}}$ to  $c_{\text{target}} = E_{pk_{\text{SI-FEIP}}}(x_\text{target})$.
Even though the \textit{server} has the original data $x_{\text{target}}$, it is not able to infer whether $x_{\text{target}}$ is in the training dataset nor not, because the generated ciphertext of $c_{\text{server}} = E_{pk_{\text{SI-FEIP}}}(x_\text{target})$ by the \textit{server} is indistinguishable from the ciphertext $c_{\text{target}}$.

\subsection{Privacy Analysis}
\textit{NN-EMD} also ensures the privacy of the output of the secure computation protocols.
Here, we present two types of inference attacks launched by the \textbf{\textit{honest-but-curious server}}.

\noindent\textbf{Inference Type I}.
Our proposed \textit{S2PHC} and \textit{S2PVC} protocols adopt the functional encryption as the underlying cryptosystems.
For both functions $f_{\text{SIIP}}(\pmb{x}, \pmb{w})$ and $f_{\text{MIIP}}((\pmb{x}_{1}, ..., \pmb{x}_{n}), \pmb{w})$ as described in Section \ref{sec:fe}, the \textit{server} is able to acquire the decryption results (i.e., the output of the first layer in NN), and the weights of the first layer (i.e., $\pmb{w}$).
The security of functional encryption scheme can ensure that the \textit{server} cannot break/infer the input $\pmb{x}$ or $(\pmb{x}_{1}, ..., \pmb{x}_{n})$.
However, an inference attack may be possible by iteratively employing FE on a specific $\pmb{x}$.
Consider the iterative training such that the \textit{curious server} may be able to collect enough polynomial equations for a specific training sample.
For instance, suppose we have one training data sample $\pmb{x}$.
For each iteration $i$ in the training phase, the \textit{server} is able to acquire $f_i = \langle\pmb{x},\pmb{w}_i\rangle$, where $f_i$ and $\pmb{w}_i$ are available or visible to the \textit{server}.
Obviously, with enough pairs of $(\pmb{w}_i, f_i)$, the \textit{server} is able to solve the linear equation system $\{f_i = \langle\pmb{x},\pmb{w}_i\rangle\}$ and acquire $\pmb{x}$.
Formally, suppose that the sample $\pmb{x}$ has $n_{\text{feature}}$ features, i.e., $\pmb{x} = (x_1, x_2, ..., n_{\text{feature}})$, and each sample is used once in one training epoch.
Let the total number of training epoch be $n_{\text{epoch}}$, and the number of periodical shuffle operations is $n_{\text{shuffle}}$.
We have the following Lemma:
\newtheorem{theorem}{Lemma}
\begin{theorem}
    \label{theo:it1}
    \textit{NN-EMD} is able to prevent \textit{Inference Type I}, if $\frac{n_{\text{epoch}}}{n_{\text{shuffle}}} < n_{\text{feature}}$
\end{theorem}
\begin{proof}
Suppose that the \textit{curious server} has advantage $\epsilon$ to infer $\pmb{x}$, which indicates it has $\epsilon$ advantage to solve the system of linear equation problems $\{f_i = \langle\pmb{x},\pmb{w}_i\rangle\}$ with determined solution.
According to theorem of PSSLS in linear algebra \cite{beezer2008first}, the \textit{curious server} has the advantage $\epsilon$ to collect $n_{\epsilon}$ linear equations for the specific sample $\pmb{x}$, where $n_{\epsilon} \ge n_{\text{feature}}$.

However, in \textit{NN-EMD}, the \textit{server} has non-negligible advantage to distinguish the ciphertext of $\pmb{x}$ among all encrypted training samples as proved in \cite{abdalla2015simple, abdalla2018multi}.
After encrypted sample shuffle by the \textit{data source}, the \textit{server} also has non-negligible advantage to learn the position of $\pmb{x}$ in the training set. 
Thus, the \textit{server} only has the advantage to collect $n_{\epsilon} = \frac{n_{\text{epoch}}}{n_{\text{shuffle}}}$ linear equations.
Here, $\frac{n_{\text{epoch}}}{n_{\text{shuffle}}} < n_{\text{feature}}$ in \textit{NN-EMD} is subject to the requirement of PSSLS theorem, namely, $n_{\epsilon} < n_{\text{feature}}$.
As a result, the \textit{curious server} has no advantage to infer $\pmb{x}$.
\end{proof}


\noindent\textbf{Inference Type II}.
The \textit{curious server} could also launch another type of inference attack by specifying ``malicious'' $\pmb{w}$ to acquire the functional private key.
For instance, by specifying $\pmb{w}=(1,0,...,0)$, the decryption result of $\langle\pmb{x},\pmb{w}\rangle$ will disclose the first element $x_1$ of $\pmb{x}$.
To prevent such an attack, we have introduced \textit{inference weights filter} into the \textit{TPA}.
Specifically, the filter module will check the vector $\pmb{w}=(1,0,...,0)$ to ensure that the number of non-zero elements is greater than a threshold $\tau$, basically, $\tau\ge 2$.
As a result, it is impossible to launch the above inference attack.

\section{Evaluation}
\label{sec:evaluation}
We evaluate the following aspects of \textit{NN-EMD}:

\noindent (\romannumeral1) To present the efficiency advantage of training time of our \textit{NN-EMD} framework, we compare its training time with that of only those closely related solutions proposed in \cite{nandakumar2019towards, xu2019cryptonn}. 
We also explore the impact of network architecture and the number of network layers in the training time in our \textit{NN-EMD} framework. 
    
\noindent (\romannumeral2) With respect to the trained model accuracy, we compare our \textit{NN-EMD} framework in a horizontal partitioning based training setting and a vertically partitioned based training setting with a baseline model, namely, a normal neural network without any privacy-preserving settings.

\noindent (\romannumeral3) As the underlying cryptosystems only work on the integer field, while the training of neural networks model works on the floating-point number field, we try to evaluate the impact of the precision on the model performance after the numeric encoding/decoding.

Note that the impact of data distribution such as non-iid and imbalanced data is beyond the scope of \textit{NN-EMD} because our framework only provides secure computation features into existing the DNN model rather than modifying the intrinsic properties of the underlying DNN model such as the network architecture.

\subsection{Experimental Setup}
To benchmark the performance of the \textit{NN-EMD} framework, we train a model of a neural network with the same topology as the one used in \cite{nandakumar2019towards} on the publicly available MNIST dataset of handwritten digits\cite{yann2010mnist} that includes 60000 training samples and 10000 test samples.
In our evaluation, each sample ($28\times 28$ image) in the MNIST dataset is mapped to a vector with a length of 784.
Besides, we also explore the framework performance on different neural network architectures and different numbers of network layers.
Essentially, we run the experiments for 5 data sources forming the \textit{client pool}.
Each data source is randomly assigned $60000/5=12000$ data samples from the MNIST dataset for the horizontal partitioning based training, while in the vertical partitioning based training, each data source is assigned 60000 data samples but only around $784/5\approx 157$ features for each sample.
Note that such a vertical setting over the MNIST data is only for illustration purposes.
We use comparable settings when evaluating the impact of the number of data sources on the model performance.
In all the experiments we utilize the same model hyperparameters of a neural network model such as learning rate, l2 regularization parameter, etc. 

\noindent\textbf{Implementation Consideration}.
We have implemented the \textit{NN-EMD} framework based on the \textit{NumPy} library to use the high-level mathematical functions in \textit{Python} programming language.
The underlying cryptosystems, namely, the functional encryption schemes, are also implemented in Python based on the \textit{gmpy2} library, which is a C-coded Python extension module that supports multiple-precision arithmetic and relies on the GNU multiple precision arithmetic (GMP) library.

In contrast to the implementation of functional encryption in \cite{xu2019cryptonn}, we incorporate the acceleration techniques used in \cite{xu2019hybridalpha} in the proposed work. 
By tracking the time cost of each decryption step in the functional encryption scheme, we find that the most inefficient computing step is the final step that computes the discrete logarithm of a small integer.
To be specific, it involves computing $f$ in $h=g^{f}$, where $h$, and $g$ are big integers while $f$ is a small integer.
To accelerate such discrete logarithm computations, we employ a \textit{bounded-table-lookup} method by initially setting up a hash table to store pair $(g,f)$ with a specified public key parameter $g$ and a positive bound $f_{b}$ where $-f_b \le f \le f_b$. 
The size of the hash table depends on the allowed encoding precision on encryption over the floating-point numbers.
Then, the final discrete logarithm  computation is a table look-up operation with complexity $\mathcal{O}(1)$, which is better compared to traditional \textit{baby-step giant-step} algorithm that has complexity $\mathcal{O}(n^{\frac{1}{2}})$.

\noindent\textbf{Environment Setup}.
All the experiments have been performed on two test platforms: \textit{\textbf{Test Platform I (TP I)}} that is a local Macbook Pro with 2.3GHz Intel Core i9 8-Core CPU and 32GB RAM, and \textit{\textbf{Test Platform II (TP II)}} that is a remote cloud service, i.e., AWS \textit{m5d.8xlarge} instance with 2.5GHz Intel Xeon 8124M 32 vCPUs and 128GB RAM.
For the evaluations of  model performance, where the client pool and the server are put on the same platform, we repeat the experiments in both the test platforms.
To simulate real scenarios, we use \textit{TP I} as the \textit{client pool} and \textit{TP II} as the \textit{server}.

\subsection{Experimental Results}

\subsubsection{Comparison with Contracted Frameworks}

\begin{table*}[t]
    \centering
    \begin{threeparttable}
    \caption{Comparison of time cost for training one mini-batch (60 samples)}
    \label{table:cmp_time_cost}
    \begin{tabular}{llllll}
        \toprule
        Proposed work & Network architecture & CPU & Threads & Mem & Training time  \\
        \midrule
        Nandakumar et al. \cite{nandakumar2019towards} & $784\mapsto128\mapsto32\mapsto10$ & 2.3GHz Intel Xeon E5-2698v3 16-Core & 1 & 250GB & $\approx 1.5$days \\
        Nandakumar et al. \cite{nandakumar2019towards} & $64\mapsto32\mapsto16\mapsto10$ & 2.3GHz Intel Xeon E5-2698v3 16-Core & 1 & 250GB & 9h24m \\
        Nandakumar et al. \cite{nandakumar2019towards} & $64\mapsto32\mapsto16\mapsto10$ & 2.3GHz Intel Xeon E5-2698v3 16-Core & 30 & 250GB & 40m \\
        CryptoNN \cite{xu2019cryptonn} & $784\mapsto128\mapsto32\mapsto10$ & 2.3GHz Intel Core i7 8-Core & 1 & 16GB & $\approx 2$days \\
        CryptoNN \cite{xu2019cryptonn} & $784\mapsto128\mapsto32\mapsto10$ & 2.3GHz Intel Core i7 8-Core & 8 & 16GB & $\approx 94$m \\
        NN-EMD (HPT) & $784\mapsto128\mapsto32\mapsto10$ & 2.3GHz Intel Core i9 8-Core & 1 & 32GB & \textbf{49.83s} \\
        NN-EMD (VPT) & $784\mapsto128\mapsto32\mapsto10$ & 2.3GHz Intel Core i9 8-Core & 1 & 32GB & \textbf{31.71s} \\
        NN-EMD (HPT) & $784\mapsto128\mapsto32\mapsto10$ & 2.5GHz Intel Xeon 8124M 32 vCPUs & 1 & 128GB & \textbf{55.63s} \\
        NN-EMD (VPT) & $784\mapsto128\mapsto32\mapsto10$ & 2.5GHz Intel Xeon 8124M 32 vCPUs & 1 & 128GB & \textbf{33.67s} \\
        \bottomrule
    \end{tabular} 
    \end{threeparttable}
    
\end{table*}

As shown in \tablename\;\ref{table:cmp_time_cost}, we compare the training time of our \textit{NN-EMD} framework with the approaches proposed in \cite{nandakumar2019towards, xu2019cryptonn}.
Note that as the codes and experimental platforms for work in \cite{nandakumar2019towards} are not publicly available, we report the experimental results reported in \cite{nandakumar2019towards} directly.
We also include the test environment reported in their papers.
In our evaluation, we use comparable experimental platforms used in \cite{nandakumar2019towards, xu2019cryptonn}, and train the model on the same MNIST dataset with the same neural network architecture.

We evaluate the \textit{NN-EMD} framework both in horizontal partitioning based training (HPT) and vertical partitioning based training (VPT) settings.
Our experimental results show that the training time of one mini-batch including 60 samples in our \textit{NN-EMD} only needs \textit{49.83} seconds and \textit{55.63} seconds in \textit{TP I} and \textit{TP II} environments, respectively.
Compared to the existing best result (i.e., 40 minutes) as reported in \cite{nandakumar2019towards} where each training sample is extracted from $28\times 28$ to $8\times 8$ to reduce the input size and the multithreaded parallelism technique is employed in the training phase, our proposed \textit{NN-EMD} reduces the training time by approximately 97.7\%.

In contrast to the approach proposed in \cite{nandakumar2019towards}, where computation at each layer of neural networks is over encrypted data, in essence, the computation over the encrypted data only occurs at one layer in \textit{NN-EMD}. This is the main reason behind the significantly lower training time in \textit{NN-EMD} compared to that of \cite{nandakumar2019towards}. 
In particular, in \textit{NN-EMD}, only the input layer is protected in our illustration and evaluation, while the rest of the parameters are still plaintext to the server. Thus, to some extent, low-level features (e.g., the output of the second layer) may reveal partial private information such as through model inversion attack as illustrated in \cite{fredrikson2015model}.
One approach to prevent such potential privacy leakage is to protect more layers of neural networks using \textit{SplitNN} technique \cite{vepakomma2018split} in our framework; this will impact training time depending on the architecture of the DNN and the splitting layers; we analyze such integration in Section~\ref{sec:evaluation:res:split}.

\subsubsection{Impact of NN Architectures and Number of Layers}

\begin{table*}[t]
    \centering
    \begin{threeparttable}
    \caption{Training time cost of one mini-batch of different network architectures in NN-EMD}
    \label{table:cmp_time_diff_networks}
    \begin{tabular}{llcc}
        \toprule
        NN Framework & Network architecture & Training time @ TP I & Training time @ TP II\\
        \midrule
        NN-Normal (\textbf{Baseline}) & $784\mapsto256\mapsto10$ & 0.00718s & 0.01169s\\
        NN-Normal (\textbf{Baseline}) & $784\mapsto256\mapsto128\mapsto64\mapsto10$ & 0.00708s & 0.01088s\\
        NN-Normal (\textbf{Baseline}) & $784\mapsto256\mapsto128\mapsto64\mapsto32\mapsto16\mapsto10$ & 0.00741s & 0.01083s\\
        \hline
        NN-EMD(HPT) & $784\mapsto256\mapsto10$ & 91.45s & 111.50s\\
        NN-EMD(HPT) & $784\mapsto256\mapsto128\mapsto64\mapsto10$ & 90.54s & 111.28s\\
        NN-EMD(HPT) & $784\mapsto256\mapsto128\mapsto64\mapsto32\mapsto16\mapsto10$ & 89.66s & 111.13s \\
        \hline
        NN-EMD(VPT) & $784\mapsto256\mapsto10$  & 55.58s & 67.48s\\
        NN-EMD(VPT) & $784\mapsto256\mapsto128\mapsto64\mapsto10$ & 55.21s & 67.26s\\
        NN-EMD(VPT) & $784\mapsto256\mapsto128\mapsto64\mapsto32\mapsto16\mapsto10$ & 56.19s & 67.05s\\
        \hline
        NN-EMD(HPT)-SplitNN & $784\mapsto256$ $|\text{split}|$ $128\mapsto64\mapsto32\mapsto16\mapsto10$ & 49.4s & 61.61s\\
        NN-EMD(HPT)-SplitNN & $784\mapsto256\mapsto128$ $|\text{split}|$ $64\mapsto32\mapsto16\mapsto10$ & 23.21s & 28.6s\\
        NN-EMD(HPT)-SplitNN & $784\mapsto256\mapsto128\mapsto64$ $|\text{split}|$ $32\mapsto16\mapsto10$ & 11.49s & 14.28s\\
        NN-EMD(HPT)-SplitNN & $784\mapsto256\mapsto128\mapsto64\mapsto32$ $|\text{split}|$ $16\mapsto10$ & 5.92s & 7.35s\\
        \hline
        NN-EMD(VPT)-SplitNN & $784\mapsto256$ $|\text{split}|$ $128\mapsto64\mapsto32\mapsto16\mapsto10$ & 32.08s & 38.35s\\
        NN-EMD(VPT)-SplitNN & $784\mapsto256\mapsto128$ $|\text{split}|$ $64\mapsto32\mapsto16\mapsto10$ & 14.97s & 17.78s\\
        NN-EMD(VPT)-SplitNN & $784\mapsto256\mapsto128\mapsto64$ $|\text{split}|$ $32\mapsto16\mapsto10$ & 7.37s & 8.74s\\
        NN-EMD(VPT)-SplitNN & $784\mapsto256\mapsto128\mapsto64\mapsto32$ $|\text{split}|$ $16\mapsto10$ & 3.79s & 4.49s\\
        \bottomrule
    \end{tabular} 
    \begin{tablenotes}
        \footnotesize
        \item[$\dagger$] HPT indicates horizontally partitioned based training (HPT) setting, while VPT represents vertically partitioned based training setting. In both of HPT and VPT, there are 5 data sources. Each mini-batch includes \textit{60 samples}. In test cases of integration with SplitNN, the blue layers occurs at the data sources while the red layers computed in the server.
    \end{tablenotes}
    \end{threeparttable}
\end{table*}

As reported in \tablename\;\ref{table:cmp_time_cost}, the training times of existing solutions such as the framework proposed in \cite{nandakumar2019towards} increases significantly as the network architecture changes.
To evaluate the impact of network architectures on the training time in our \textit{NN-EMD} framework, we train DNN models with different architectures on the MNIST dataset with the same number of data sources.
As presented in \tablename\;\ref{table:cmp_time_diff_networks}, the training time for our proposed approach is only impacted by the number of nodes in the first hidden layer.
When the network architecture of the rest of the layers changes, the training time does not change compared to the normal neural networks without a privacy-preserving setting.

For further verification of such a claim, we conducted additional experiments with a large number of hidden layers.
As shown in \figurename\;\ref{fig:cost_time_diff_layers}, we measure the training time of one mini-batch in \textit{NN-EMD} in different training settings (i.e., HPT and VPT) and vary the number of hidden layers from 1 to 30 where each layer includes 64 neural nodes.
As can be seen, the training time does not change drastically as in other existing solutions as the number of hidden layers increases.

\subsubsection{Evaluation of Accuracy}

\begin{figure*}[!t] 
	\centering 
	\subfloat[{Model accuracy with precision=4}]{
		\includegraphics[scale=0.33]{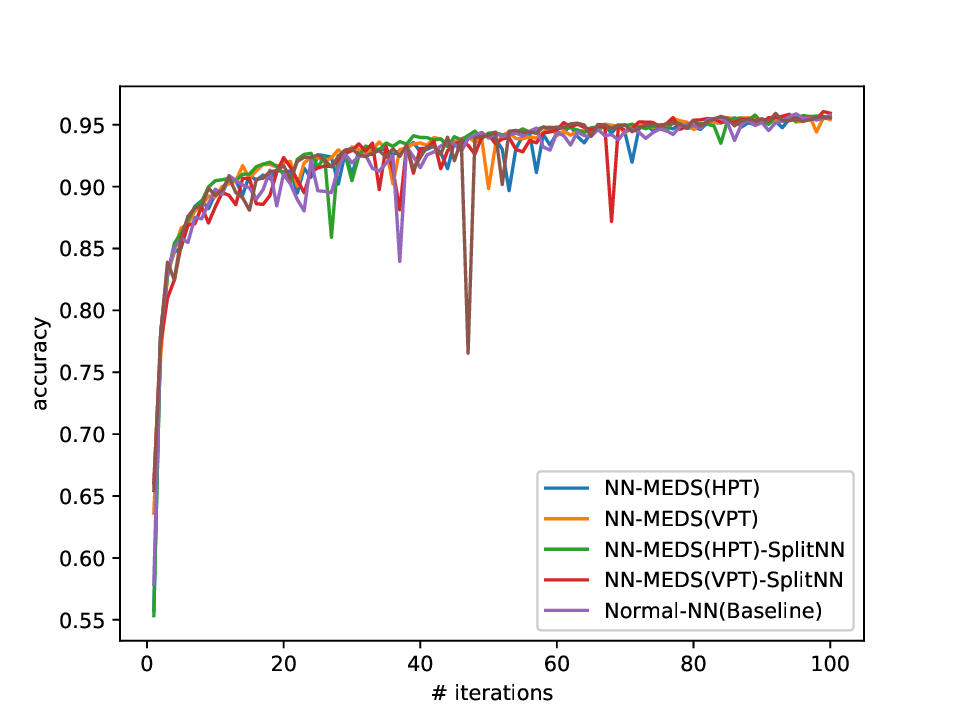}
		\label{fig:acc:cmp}
		
	}
	\hfil
	\subfloat[Impact of precision on the model accuracy]{
		\includegraphics[scale=0.33]{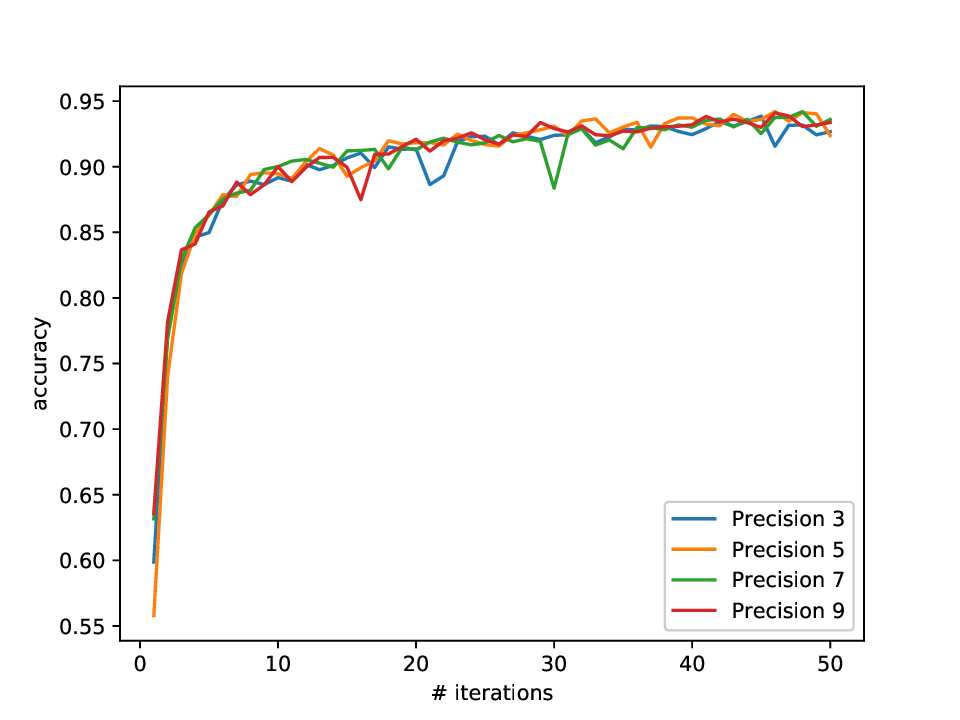}
		\label{fig:acc:diffp}
		
	}
	\subfloat[Training time of one mini-batch]{
		\includegraphics[scale=0.33]{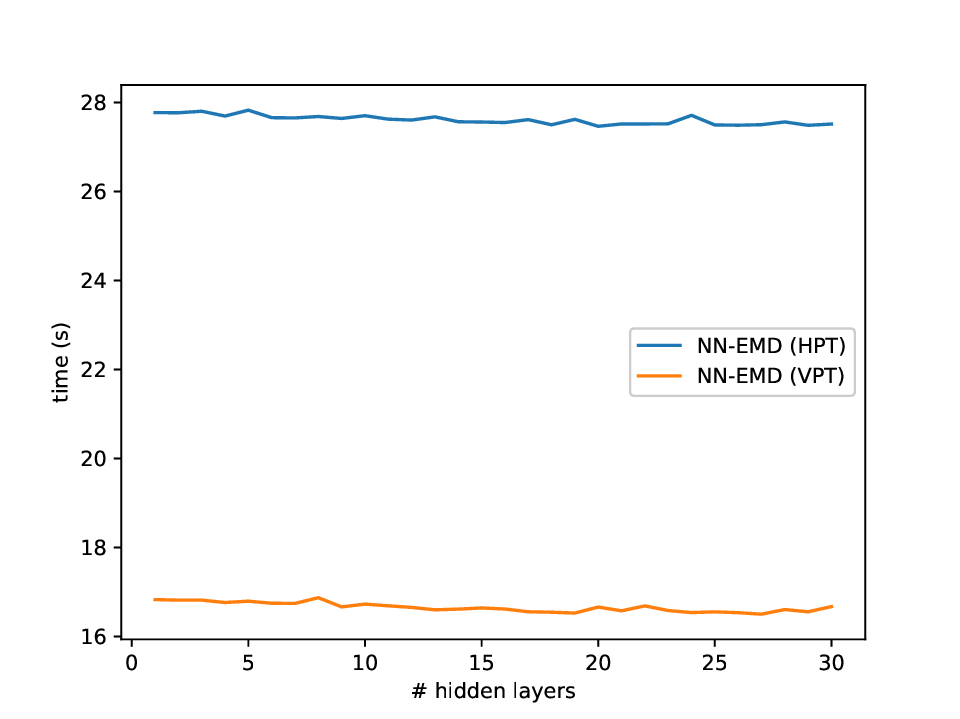}
		\label{fig:cost_time_diff_layers}
		
	}

	\caption{Comparison of model accuracy and training time of one mini-batch as number of hidden layers increase. Note that the network architecture used for model accuracy comparison in \textit{Normal-NN} and \textit{NN-EMD} is $784\mapsto512\mapsto256\mapsto128\mapsto64\mapsto32\mapsto10$. Each hidden layer of network architecture in right figure includes 64 neural nodes and the results are generated on the \textit{TP II} platform.}
	\label{fig:acc} 
\end{figure*}

Except for the performance with respect to the training time, we compare our framework with a baseline neural network framework (Normal-NN) that has the same network architecture but without any settings of privacy-preserving approaches.
As shown in \figurename\;\ref{fig:acc:cmp},  our proposed \textit{NN-EMD} framework can achieve comparable model accuracy compared to a normal neural network both in HPT  and VPT. 
Further, the results in \figurename\;\ref{fig:acc:diffp} shows that the precision setting does not have an effect on the model accuracy.

\subsubsection{Integration with SplitNN}
\label{sec:evaluation:res:split}
To explore the trade-off between privacy and training time efficiency, we also integrate \textit{SplitNN} method \cite{vepakomma2018split} into our \textit{NN-EMD}.
Specifically, we adopt the network architecture of $784\mapsto256\mapsto128\mapsto64\mapsto32\mapsto16\mapsto10$ and split the neural network layers at various points in both HPT and VPT settings, in which usually more layers computed locally indicate higher privacy guarantee, as analyzed in \cite{vepakomma2018split}.
As reported in \tablename\;\ref{table:cmp_time_diff_networks}, the training time depends on the number of neural nodes at the first layer computed at the server after the splitting of the network architecture.
Thus, the impact of the integration of \textit{SplitNN} method on the training efficiency depends on how to split the network architecture and what type of network architecture is used to train the DNN model, since proposed secure computation in \textit{NN-EMD} only occurs at the splitting layer.
Furthermore, as shown in \figurename~\ref{fig:acc:cmp}, the integration does not impact the model accuracy.

\subsubsection{Deployment in Client-Server Scenario}

\begin{table}[t]  
    \centering
    \begin{threeparttable}
    \caption{Time cost for different data source \# in client-server setting in NN-EMD}
    \label{table:cmp_time_diff_ds}
    \begin{tabular}{cccc}
        \toprule
        & & \multicolumn{1}{c}{Pre-process Time} & \multicolumn{1}{c}{Training Time} \\
        \cmidrule(ll){3-3} \cmidrule(ll){4-4}
        Training & Sources \# & Client (TP I) &  Remote AWS (TP II)\\
        \midrule
        HPT Setting & 5  & 1.2769s & 56.02s\\
        HPT Setting & 7  & 1.0579s & 55.95s\\
        HPT Setting & 9  & 1.1195s & 55.86s\\
        HPT Setting & 11 & 1.1246s & 55.87s\\
        HPT Setting & 13 & 1.1463s & 56.04s\\
        HPT Setting & 15 & 1.1260s & 56.13s\\
        \hline
        VPT Setting & 5  & 0.3716s & 33.75s\\
        VPT Setting & 7  & 0.3363s & 32.48s\\
        VPT Setting & 9  & 0.2372s & 31.93s\\
        VPT Setting & 11 & 0.2214s & 31.50s\\
        VPT Setting & 13 & 0.1867s & 31.24s\\
        VPT Setting & 15 & 0.1711s & 31.17s\\
        \bottomrule
    \end{tabular} 
    \begin{tablenotes}
        \footnotesize
        \item[$\dagger$] The neural networks architecture used in this experiment is $784\mapsto128\mapsto32\mapsto10$. Note that the cost time reported here is for only one mini-batch that includes 60 samples.
    \end{tablenotes}
    \end{threeparttable}
\end{table}

To evaluate the impact of the number of data sources on the training time, we have deployed our end-to-end \textit{NN-EMD} system in a \textit{client-server} scenario.
In this experiment, our local machine (\textit{TP I}) plays the role of \textit{client pool} with varying number of data sources to pre-process the encrypted training datasets, while the remote AWS instance (\textit{TP II}) plays the role of the \textit{server} to train the neural network model based on these encrypted data samples.

As shown in \tablename\;\ref{table:cmp_time_diff_ds}, we present the training time for both the \textit{client pool} and the \textit{server}.
All reported times for the \textit{server} side is based on one mini-batch, while the time reported for the \textit{client pool} is for one mini-batch per data source. 
In the case of the horizontal partitioning based training, the training time of \textit{NN-EMD} framework does not change drastically like existing solutions as the number of data sources increases.
In the case of the vertical partitioning based training, each data source pre-processes the same number of data samples.
As the total number of features is fixed, the number of features from each data source decreases as the number of data sources increase, and hence the pre-processing time decreases, while the training time still does not change drastically.

\section{Related Work}
\label{sec:related_work}

\noindent\textit{Secure Computation}.
Secure computation, also known as secure multi-party computation (SMC) or multi-party computation (MPC), has shown its promise in supporting computational tasks for applications that process privacy-sensitive data since the first formal secure two-party computation (2PC) was proposed by Yao et al. \cite{yao1982protocols}. 
Even though the secure computation has been explored for around 40 years, the practical deployment of secure computation solutions is still a challenge, especially, in the era of big data.

Generally, there are two research directions towards achieving the goal of secure computation, namely, constructing general-purpose SMC, or proposing special-purpose SMC.
To construct general purpose SMC, existing solutions can be of two categories: (\romannumeral1) protocols such as those proposed in \cite{huang2011faster, wang2017global, wang2017authenticated, chase2019reusable} that are usually built on the \textit{garbled circuits} \cite{bellare2012foundations} and \textit{oblivious transfer} techniques \cite{asharov2017more}, and (\romannumeral2) protocols such as those proposed in \cite{lopez2012fly, damgaard2012multiparty, baum2016better} that are based on the homomorphic cryptosystems \cite{gentry2009fully} (e.g., fully homomorphic encryption).
However, these two kinds of SMC solutions have limitations with regards to either the large volumes of ciphertexts that need to be transferred or the inefficiency of computation involved (i.e., unacceptable computational time).
To provide more applicable secure computation solutions, several special-purpose SMC approaches such as those proposed in \cite{canetti1996adaptively, keller2018overdrive, araki2018choose} have been proposed to address special computational tasks such as additive functions, as the general computational tasks are not required in most of the application scenarios.
In this paper, our proposed secure computation protocols are also essentially special-purpose SMCs that only address the computation of inner-products and matrix multiplications.

\noindent\textit{Computable Ciphertext}.
As we discussed above, a branch of secure computation approaches is based on homomorphic encryption schemes.
Homomorphic encryption is a form of cryptosystems that allows computation over ciphertexts, where the processed result is still in ciphertext, but the decryption of that ciphertext matches the result of the computation performed on the corresponding plaintexts.
In general, fully homomorphic encryption \cite{lewko2010fully} is able to achieve a general purpose SMC, while partially homomorphic encryption such as additive homomorphic encryption \cite{paillier1999public, damgaard2001generalisation} can construct additive secure computation protocols.
Existing homomorphic encryption schemes are still not efficient enough, especially, when applying to the large-scale computational tasks.
On the other hand, the recently proposed functional encryption approach \cite{lewko2010fully, boneh2011functional} shows another promising direction to achieve the task of computation over a ciphertext.
For instance, to construct functional encryption schemes for general functionality, most of the recently proposed approaches such in \cite{goldwasser2014multi,boneh2015semantically,waters2015punctured,garg2016candidate,carmer20175gen, lewi20165gen} focus on the theoretical feasibility or functionality existence.
Unlike homomorphic encryption schemes, functional encryption schemes rely on a trusted third party authority to provide private key service.
In this paper, we have adopted the functional encryption schemes that are only applicable to inner-products \cite{abdalla2015simple, abdalla2018multi}.

\noindent\textit{Privacy-preserving Machine Learning.}
The goal of privacy-preserving machine learning is to protect the privacy of training data while still generating a well-trained model.
To achieve that goal, most of the popular techniques such as differential privacy \cite{abadi2016deep}, federated learning \cite{shokri2015privacy}, general-purpose SMC protocols\cite{mohassel2017secureml, rouhani2018deepsecure} and computable cryptosystems \cite{xu2019cryptonn, nandakumar2019towards, gilad2016cryptonets, hesamifard2017cryptodl, jiang2018secure} have been applied in different machine learning models.
Except for the approaches proposed in \cite{xu2019cryptonn, nandakumar2019towards}, all existing cryptosystem based privacy-preserving machine learning approaches either focus on the simple traditional machine learning models such as a linear regression model \cite{nikolaenko2013privacy} and a logistic regression model \cite{aono2016scalable} or only work on the inference phase of the neural network model \cite{ gilad2016cryptonets, hesamifard2017cryptodl, jiang2018secure}.
It is still a huge challenge to train a DNN model over encrypted data, especially, on encrypted training dataset composed of data from multiple sources.

\section{Conclusion}
\label{sec:conclusion}
Training neural network models over encrypted data show significant promise towards addressing strong privacy requirements of both the users and regulations, while taking full advantage of an existing ML platform as a service infrastructure.
However, there is a lack of efficient and practical privacy-preserving solutions for training a neural network based ML system over privacy-sensitive datasets.
We have proposed \textit{NN-EMD}, a novel neural network framework that supports training a neural network model on a dataset where the data is composed, both horizontally and vertically, of encrypted datasets from multiple data sources.
Our evaluation shows that \textit{NN-EMD} can reduce the training time by 97\% while still providing the same model accuracy and strong privacy guarantee as compared to most of the recent comparable approaches.
Furthermore, the depth and complexity of neural networks do not affect the training time despite introducing a privacy-preserving \textit{NN-EMD} setting.
Future work includes applying the \textit{NN-EMD} framework in a complex edge computing environment.

\section*{Acknowledgment}
This work was performed while James Joshi was serving as a Program Director at NSF; and the work represents the views of the authors and not that of NSF.
Chao Li acknowledges the partial support by Fundamental Research Funds for the Central Universities (No. 2019RC038).

\balance

\bibliographystyle{IEEEtran} 
\bibliography{reference.bib}

\end{document}